\RequirePackage{fix-cm}
\documentclass{svjour3}       
\smartqed  
\usepackage{graphicx}
\usepackage{natbib}
\usepackage{amsmath}
\begin{document}

\title{Clustering Activity-Travel Behavior Time Series using Topological Data Analysis
}

\author{Renjie Chen*         \and
        Jingyue Zhang \and
        Nalini Ravishanker\and
        Karthik Konduri 
}


\institute{Renjie Chen* \at
              Department of Statistics, University of Connecticut \\
              Tel.: +860-450-6660\\
              \email{renjie.chen@uconn.edu}           
           \and
           Jingyue Zhang \at
              Department of Civil and Environmental Engineering, University of Connecticut\\
              \email{jingyue.zhang@uconn.edu}
              \and
              Nalini Ravishanker\at
              Department of Statistics, University of Connecticut \\
              \email{nalini.ravishanker@uconn.edu}
          \and
          Karthik Konduri \at
          Department of Civil and Environmental Engineering, University of Connecticut\\
          \email{karthik.konduri@uconn.edu}
}

\date{Received: date / Accepted: date}

\maketitle

\begin{abstract}
Over the last few years, traffic data has been exploding and the transportation discipline has entered the era of big data. It brings out new opportunities for doing data-driven analysis, but it also challenges traditional analytic methods. 
This paper proposes a new Divide and Combine based approach to do K-means clustering on activity-travel behavior time series using features that are derived using tools in Time Series Analysis and  Topological Data Analysis. 
Our approach facilitates a case study, where each individual’s daily activity-travel behavior is characterized as a categorical time series consisting of three different levels. Clustering data from five waves of the National Household Travel Survey ranging from 1990 to 2017 suggests that activity-travel patterns of individuals over the last three decades can be grouped into three clusters. Results also provide evidence in support of recent claims about differences in activity-travel patterns of different survey cohorts. The proposed method is generally applicable and is not limited only to activity-travel behavior analysis in transportation studies. Driving behavior, travel mode choice, household vehicle ownership, when being characterized as categorical time series, can all be analyzed using the proposed method.  
\end{abstract}

\section{Introduction}

Transportation data is exploding in recent years owing to the improved technologies for data collection and storage. A vast amount of data are generated and collected for various purposes. Examples include smartcard data collected by transit operators, mobile phone traces collected by phone carriers, traffic data collected via sensors, smart cameras, global positioning system (GPS) data by road operators, and user’s Wi-Fi locations collected by internet providers. 
There is an increasing number of studies attempting to leverage big data for answering different transportation-related questions.  Studies have sought to use big data for improving traffic management. For example, \cite{silva2015} proposed to use data collected by the drivers using apps like Waze and Google Maps 
to improve urban mobility. \cite{Figueiras2016} proposed to aggregate big data from various sources for implementing dynamic tolling to reduce traffic congestion. Other studies used big data for revealing individuals’ mobility patterns \citep{CALABRESE2013301,Candia_2008,Kwan2000,Huang12710}. For example, \cite{Candia_2008} used mobile phone data with time and space resolution to explore collective behavior and detect anomalous events of human activity patterns. \cite{Huang12710} used 7-year transit smartcard data to reveal commute patterns and explore the relationship of job and housing locations of travelers in Beijing, China. When referring to big data analysis, current studies only focus on passively collected data (i.e., phone trace data, smartcard data and sensors data). However, such data has limitations: 1) the datasets do not include the socioeconomic and demographic information of individuals, which are important for understanding the underlying behavior mechanism of individuals’ activity-travel behaviors; 2) the data is not carefully collected to represent a random sample of the population; 3) the data usually requires intensive processing before being used for analysis \citep{CALABRESE2013301}. 
On the other side of the spectrum we have traditional surveys that overcome these limitations.
Due to the high expense of conducting surveys, most surveys only collect data from a small sample within limited temporal and spatial scales. 
However, the National Household Travel Survey increased in recent years and it is the largest travel survey that collects detailed trip information. 
As aforementioned, actively collected survey data shows advantages for analyzing activity-travel patterns. It not only contains activity-travel behavior of each individual, but also includes socioeconomic and demographic information for revealing the underlying mechanisms of the behavior of individuals.

Understanding the relationship between individuals’ activity-travel behaviors and their socioeconomic and demographic characteristics can help transportation planners promote efficient solutions and policies for a given region. When analyzing activity-travel behavior using survey data, researchers tend to focus on one or two aspects of activity and travel (i.e., trip rate, mode choice, or activity type). They often ignore the temporal dimension of activity-travel behaviors (i.e., timing, duration and sequential order of activity and travel). One way to incorporate these 
is through a categorical time series characterization \citep{Wilson2001,RECKER1985279,Shoval2007,ZHANG201896,GOULIAS1999535}. Each data point of the time series represents a minute spent in either travel or activity over the course of a day.

It is useful to first cluster the categorical time series,
separating individuals into groups of distinct temporal behaviors and then explore the relationship between the temporal behaviors and their demographic characteristics.
Two types of clustering methods have been widely adopted, 
the sequence alignment method \citep{joh2001pattern,Wilson2001,RECKER1985279,Pas1988,Shoval2007, ZHANG201896} and the Markov modeling approach \citep{GOULIAS1999535}. 
The sequence alignment method was first developed in molecular biology for calculating the sequential similarity between DNA strings. The method is based on the Levenshtein distance, also called the Edit distance, which is defined as the smallest number of changes made in the elements to equalize two sequences \citep{joh2001pattern}. The method is very computationally intensive, and so it has only been applied for analyzing small datasets. 
The Markov model is also useful for characterizing categorical time series and estimates the probability of transitioning from an activity-travel model at time $t$ to another activity at time $t+1$. The Markov model is generally most suitable when the time series patterns change periodically.

We propose an approach that constructs useful   
features from time series using frequency domain properties and Topological Data Analysis (TDA)\footnote{A brief review is provided in the appendix. For details on
TDA, see \cite{harer2010,wang2018topological}}. 
Our approach then clusters the series into groups based on these features. That is, we propose a sequence alignment method based on the dissimilarity between series using TDA based features.
\textit{In order to attain computational speed in applying this approach, we propose a divide and combine scheme for the implementation.}

The rest of the paper is organized as follows.
Section \ref{features} shows how we can construct useful features of time series using TDA. In Section \ref{DCS}, we discuss K-means clustering of a large number of time series based on these features, by using a divide and combine scheme to handle the computational burden. Both Sections \ref{features} and \ref{DCS} provide generic descriptions that can be used with any set of categorical time series.
Section \ref{post} discusses this approach on a case study on diurnal activity-travel behavior of a large number of participants from the National Household Travel Survey (NHTS)/National Personal Travel Survey (NPTS). Section \ref{summary} presents a summary of our contributions and 
ideas for future research. The appendix provides a brief review of TDA and the persistence landscape construction.

\section{TDA Based Features of Categorical Time Series} \label{features}

Section \ref{features} describes feature extraction from categorical time series using TDA on their frequency domain representations. 
Let $x_{n,t}$, $t=1,\ldots,T$ and $n=1,\ldots,N$ denote a large set of $N$ categorical time series, each of length $T$ and each assuming $J$ levels.
The feature extraction from each categorical time series consists of two steps.

In the first step, we convert the time series $x_{n,t}$ to their frequency domain representations, the \textit{Walsh-Fourier transforms} (WFT), which are useful in representing ``sequency patterns'' in categorical time series \citep{Stoffer1991}. We use an efficient  algorithm 
developed by  \cite{Shanks1969} to compute the fast WFT using discrete, orthogonal Walsh functions generated by a multiplicative iteration equation. Walsh functions constitute a set of piecewise constant functions which assume a value of $-1$ or $+1$ on sub-intervals of time defined by dyadic fractions. 
Although the fast WFT captures the sequency properties of the time series, its usefulness as a feature in  clustering the $N$ time series may be mitigated when a time series has low (rather than high) sequency patterns. It is useful to retain the dominant sequency features of the WFT, while removing redundancies. 

For this purpose, in the second step of the feature construction, we convert the WFT of the time series into a first-order \textit{persistence landscape} \citep{Bubenik2015}, which is a summary statistic in 
topological data analysis (TDA) and is easy to compute and combine with tools from statistics and machine learning.  
The appendix gives a brief review of concepts in TDA, which is being increasingly explored for analyzing big, complex data \citep{wang2018topological, stolz2017persistent}, and in particular, a description of the first-order persistence landscape corresponding to a function.
The persistence landscape of the WFT will be useful to pull up the strongest temporal patterns in the categorical time series, and will be employed as features in the clustering algorithm. 
The two-step procedure is described below.
\vskip1em

\noindent \textbf{Step 2.1. Fast Walsh-Fourier Transform of a Categorical Time Series.}
Construct the fast WFT using the method of \cite{Shanks1969} to decompose the $n$th time series $x_{n,1}, \ldots, x_{n,T}$ into a sequence of Walsh functions, each representing a distinctive binary sequency pattern. If the time series length $T$ is not a power of $2$, let $T_2$ denote the next power of $2$. For example, if $T=1440$, then  $T_2 = 2^{11} =2048$. 
Use zero-padding to obtain a time series of length $T_2$, i.e., set  $x_{n,T+1}, x_{n,T+2}, \ldots, x_{n,T_2} = 0$. 

For $j=0,\ldots, T_2-1$, let $\lambda_j = j/T_{2}$ denote the $j$th sequency. 
Let $W(t, j)$ denote the $t$-th Walsh function value in sequency $\lambda_j$. Walsh functions are iteratively generated as follows \citep{Shanks1969}: 
\begin{eqnarray}
W(0,j) &=& 1, j=0, 1, \ldots, T_2-1, \notag  \\
W(1, j)&=& 
\begin{cases}
1 & j=0, 1,\ldots, (T_2)/2-1 \\
-1 &  j=(T_2)/2, (T_2)/2+1, \ldots, T_2-1
\end{cases}  \notag \\
W(t, j) &=& W([t/2],2j) \times W(t-2[t/2], j), \\
t&=&2, \ldots, T_2-1, \hspace{0.1in} j=0, 1, \ldots, T_2-1, \notag
\end{eqnarray}
where $[a]$ denotes the integer part of $a$. For more details on Walsh functions, please refer to \cite{Stoffer1991}.

The Walsh-Fourier Transform (WFT) of $x_{n,t}$ is computed as
\begin{equation}
 d_T(n,\lambda_j) = \frac{1}{\sqrt{T_2}} \sum_{t=1}^{T_2} x_{n,t}  \ W(t, j)), \hspace{0.1in} 0\leq j \leq T_2 -1.
\end{equation}
The length of $d_T(n, \lambda_j)$ is $T_2$. We use C++ code to compute the fast WFT and its computational complexity is $O(T \log (T))$  \citep{Shanks1969}.

\vskip1em
\noindent \textbf{Step 2.2. Persistence Landscape Corresponding to a WFT.}
We construct a first-order persistence landscape (see the appendix for a brief review) corresponding to the WFT $d_T(n,\lambda_j), j=0, 1, \ldots, T_2-1$ of the time series $x_{n,t}$ as follows.
Denote the minimum and maximum of the WFT values of the time series $x_{n,t}$ by 
$$
d_{n, \min}   = \min_{j} d_T (n,\lambda_j)  \mbox{ and }  d_{n, \max}= \max_{j} d_T (n,\lambda_j).
$$
Let 
$$
D_{\min} =  \min_{n} d_{n, \min}  \mbox{ and } D_{\max} = \max_{n} d_{n, \max}
$$ 
denote the minimum and maximum values of the WFTs across all $N$ time series.

We construct the first-order persistence landscape of length $L$, for a time series indexed by $n$.
Usually, $L$ is chosen to be considerably smaller than the length $T$ of the time series for computational speed, while not making it too small to make the persistence landscape from the WFT ineffective to capture essential features of the time series.
We have chosen $L=100$ based on the empirical observation that it captures the strongest temporal patterns in the activity-travel categorical time series.

The first-order persistence landscape of $x_{n,t}$ is obtained for $\ell = 1, 2,\ldots, L$ as  
\begin{equation}
\mbox{PL}(n,\ell) = \min (V_1(n,\ell), V_2(n,\ell))_{+}
\label{PL1}
\end{equation}
where
\begin{eqnarray*}
V_1(n,\ell) = D_{\min} + \frac{(\ell-1) (D_{\max}-D_{\min})}{L-1} - d_{n, \min}, \nonumber\\
V_2(n,\ell) = d_{n,\max} - D_{\min} - \frac{(\ell-1) (D_{\max} - D_{\min})}{L-1}, 
\end{eqnarray*}
and $(a)_+$ denotes the positive part of a real number $a$. 
For $\ell = 1, 2,\ldots, L$ and $n=1,\ldots,N$, the $\mbox{PL}(n, \ell)$ are piecewise linear functions that constitute features constructed for each of the $N$ time series and will be input into a clustering algorithm described in the next section.

\section{Divide and Combine K-means Clustering} \label{DCS}

We use the persistence landscapes $\mbox{PL}(n, \ell)$ for $\ell = 1, 2,\ldots, L$ and $n=1,2,\dots, N$ as features to cluster the $N$ series into homogeneous groups via the K-means algorithm. When $N$ is large,  
we can gain efficiency by operating the algorithm in parallel on multiple processors.
We use a divide and combine approach for implementing the K-means algorithm using Message Passing Interface (MPI) for parallel computing in C++. 
This significantly reduces the computing time and automatically resolves the limited memory and power restrictions of a single computer. We use the University of Connecticut (UConn) High Performance Computing (HPC) cluster with $100$ cores. The nodes consist of mixed four versions of Xeon processors (Xeon E5-2650, Xeon E5-2680 v2, Xeon E5-2690 v3, and Xeon E5-2699 v4), each having 36 cores and 156 GB; since we use $100$ cores, we would receive nodes with different configurations.
The procedure consists of several steps.

\begin{itemize}

\item[3.1.] \textbf{Data Division into $S$ Processors}. 
Denote the ordering of the categorical time series as $\Delta = (1,2$,
 $\ldots, N)$. 
We randomly divide the full data set of size $N$  categorical time series  
into $S$  
sets, so that each set consists of  $N_s$  
time series, which is a manageable number to  analyze (in parallel) on each of $S$ processors on the UConn HPC cluster.  The division is done by randomly sampling the indices of the $N$ time series without replacement and then assigning the first $N_1$ time series to the first processor, successive $N_2$ series  to the second processor, etc. Usually, we would assume that  
$N_{1} = N_{2} = N_{S-1} =[N /S]$ and assign the remaining time series to the $S$-th processor. The random sampling orders of the indices are saved into the vector $\mathbf{r}$.

\item[ 3.2.] \textbf{Feature Extraction Within Each Processor}. 

\begin{itemize}
\item [ 3.2.1.] Obtain the WFT of each categorical time series, following Step 2.1. 

\item [ 3.2.2.] Convert the WFT to a first-order persistence landscape, following Step 2.2. 
\end{itemize}

\item [ 3.3.] \textbf{K-means Algorithm on Parallel  Processors}. We implement the K-means algorithm independently on each processor $s$,  using as features the persistence landscapes of length $L$ from each time series. 
Select the number of clusters $K$. The entire algorithm will be run for different choices of $K$. 
We also set the maximum number of iterations to be $I$, chosen to be $100$. We set the iteration counter at $i=0$. 
We implement the following steps. 

\begin{itemize}
\item [ 3.3.1.] Set $i=i+1$. 
Generate centroids of each of the $K$ clusters, each of length $L$, as follows:
\begin{itemize}
\item[(i)] if $i=1$, generate the centroids for each of the $K$ clusters randomly on each processor $s$ which corresponds to $N_s$ time series. Each of the $L$ centroid components are  drawn from a
Uniform$(a_{1,s}(\ell), a_{2,s}(\ell))$ distribution, where $a_{1,s}(\ell) = \min_{n_s = 1}^{N_s} \mbox{PL}_s(n_s, \ell)$ and 
$a_{2,s}(\ell) = \max_{n_s = 1}^{N_s} \mbox{PL}_s(n_s, \ell)$.

\item[(ii)] if $1 < i \le I$, use the centroids sent by the master processor at the end of Step 3.3.3.
\end{itemize}
Run the K-means algorithm independently on each processor $s$ (note that the K-means algorithm itself includes $1,000$ iterations by default). For $s=1,\ldots,S$ and iteration $i$,  save into $\mathcal{C}_{i,s} = \{ \mathbf{c}_{i,s,1}, \ldots, \mathbf{c}_{i,s,K} \}$ the set of $L$-dimensional centroids $\mathbf{c}_{i,s,k}$ from cluster $k$, for $k=1,\ldots,K$. 
Set a flag for each processor as follows:
\begin{itemize}
\item[(i)] if $i=1$,  set a flag $f_s =1$ for each $s$.
\item[(ii)] if $2  \le i \le I$,  set $f_s =1$ if cluster labels change after the K-means algorithm on processor $s$, else set  
$f_s =0$.
\end{itemize}

\item [ 3.3.2.]  
For $s=1,\ldots, S$, processor $s$ returns to the master processor the set of centroids $\mathcal{C}_{i,s}$  
and the flag $f_s$.   For any iteration $1 \le i \le I$, 
\begin{itemize}
\item[(i)] if at least one of the $S$ flags is set at 1, the procedure of centroid selection must be iterated further; go to Step 3.3.3.

\item[(ii)] if all the flags are set at 0, the selection of centroids is complete; go to Step 3.4.
\end{itemize}

\item [ 3.3.3.] The master processor applies the same K-means algorithm with $K$ clusters on the centroids $\{ \mathbf{c}_{i,1,1}$, 
$\mathbf{c}_{i,1,2},\ldots, \mathbf{c}_{i,1,K}, \mathbf{c}_{i,2,1}, \ldots, \mathbf{c}_{i,S,1}, \ldots, \mathbf{c}_{i,S,K}\}$,  and updates the new set of centroids as $\mathcal{C}^*_{i}=\{\mathbf{c}_{i,1}^*, \mathbf{c}_{i,2}^*$, 
$\ldots,\mathbf{c}_{i,K}^*\}$. 
Note that each $\mathbf{c}_{i,s,k}$ is used an input into the K-means on centroids and $\mathcal{C}^*_{i}$ is the set of centroids after K-means. 
The master processor then sends the set $\mathcal{C}^*_{i}$ back to all $S$ processors.
For example, when $S=2$ and $K=2$, the master processor receives centroids from all $S$ processes, i.e., $\{\mathbf{c}_{i,1,1}, \mathbf{c}_{i,1,2}, \mathbf{c}_{i,2,1}, \mathbf{c}_{i,2,2}\}$, and generates the set $\mathcal{C}^*_{i}=\{\mathbf{c}_{i,1}^*, \mathbf{c}_{i,2}^*\}$ from the K-means on centroids algorithm, which is broadcast  to all $S$ processors, so that each of them may use these centroids in Step 3.3.1.
\end{itemize}

\item [ 3.4.] \textbf{Combine Results from $S$ Processors}. All $S$ processors return cluster labels $\mathbf{\ell}_s = \{l_{n_s}, n_s=1, 2, \dots, N_s\}$, where $l_{n_s}$ denotes  the cluster label for the $n_{s}$-th subject. Each processor also returns to the master processor its Within-Cluster Sum of Squares defined as 
$$
\mbox{WCSS}_s = \sum_{k,n_s, \ell} (\mbox{PL}_s(n_s, \ell) - \mathbf{c}_{s,k}(\ell))^2I(l_{n_s}=k),
$$ 
where $I(l_{n_s}=k)$ is the indicator function.
The master processor saves the cluster labels from the $S$ processors in order, $\mathcal{r}=((\mathbf{\ell}_s), s=1, 2, \ldots, S), (\mathbf{\ell}_s) = (l_1, l_2, \ldots, l_{n_s})$.
Let  
\begin{equation}
\mbox{WCSS} = \sum_{s=1}^{S} \mbox{WCSS}_s
\label{WCSS}
\end{equation}
denote the Total Within Cluster Sum of Squares.

\end{itemize}

\begin{figure*}
\includegraphics[width=\textwidth]{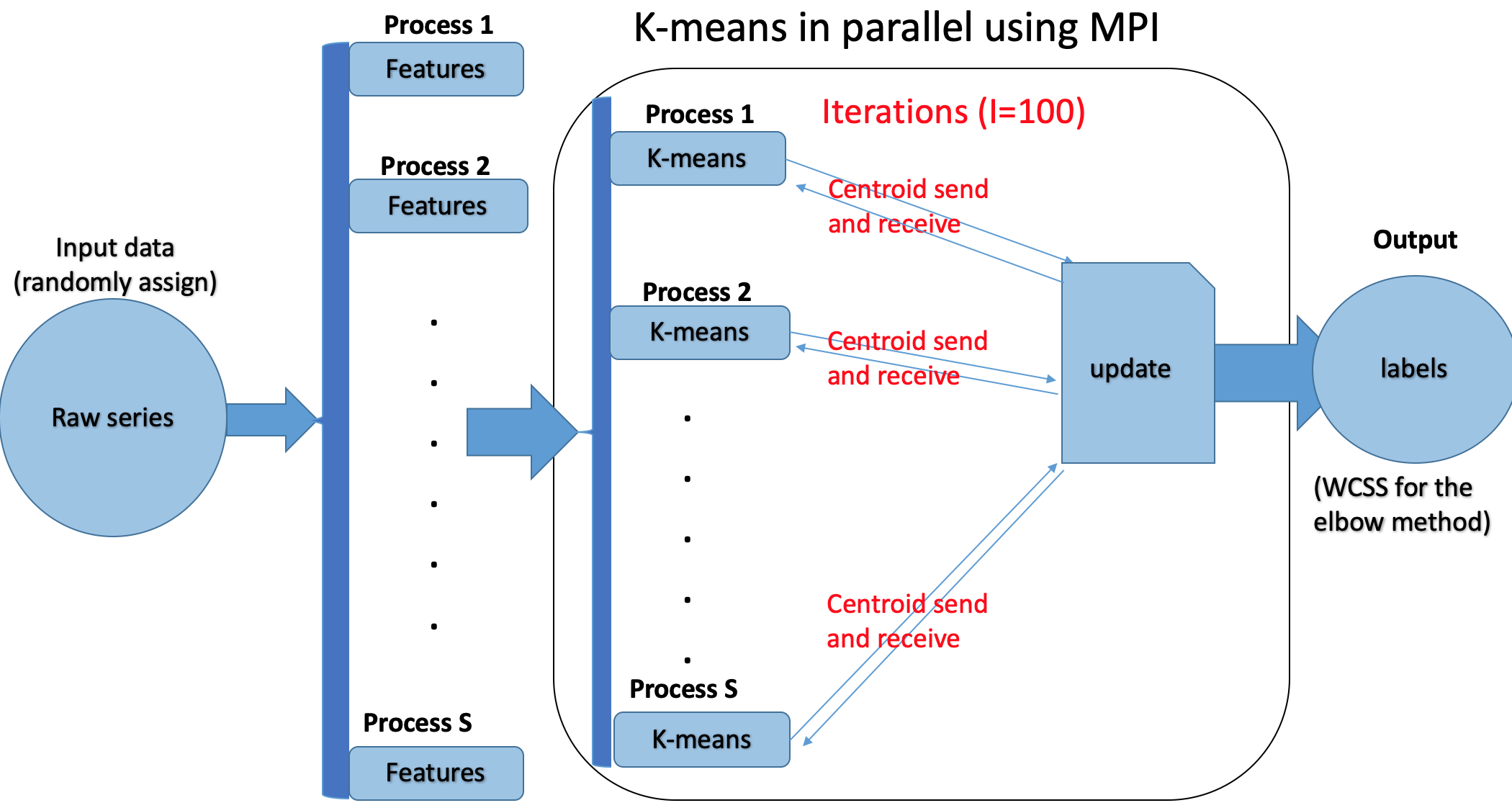}
\caption{An Overview of Implementing the Divide and Combine Scheme.}
\label{clusteringprocedure}
\end{figure*}

Figure \ref{clusteringprocedure} gives an overview of all the steps.
The final outputs from the entire procedure are: the random sampling orders $\mathbf{r}$; the WFT from each processor; the first-order persistence landscapes from each processor; the cluster labels $\mathcal{r}$; and the $\mbox{WCSS}$.  For doing interpretations by using the original time series with the cluster labels, $\mathcal{r}$, we can use $\mathbf{r}$ on the raw time series again to make the ordering $\Delta$ match with $\mathcal{r}$.

\section{Case Study: Analysis of Within-Day Activity-Travel Patterns} \label{post}

In this section, we present a detailed case study of applying our TDA based clustering procedure to activity-travel patterns from participants in multiple waves of National Household Travel Survey data ranging from 1990 to 2017. 
Following a motivation of this case study in section \ref{csm}, we provide a detailed data description in section \ref{Data} and the study design in section \ref{SDn}. 
In section \ref{respclust}, we give a discussion of the divide and conquer algorithm that uses TDA derived feature clustering described in Sections \ref{features} and \ref{DCS}. Section \ref{interpret} discusses the interpretation of results.

\subsection{Motivation of the Transportation Case Study}\label{csm}

As mentioned in the introduction, the large-scale actively collected travel survey data provides tremendous opportunities for conducting data-driven analysis for understanding activity-travel behaviors. The algorithm described in Sections \ref{features} and  \ref{DCS} is applied to identify clusters of individuals based on their intra-day activity-travel patterns. In particular, we are interested in investigating whether activity-travel behavior varies across different generation cohorts, employment status, income, or gender. These four factors have been acknowledged in the literature as strongly associated with activity-travel behavior. To this end, the primary objective of this case study is to use the proposed approach to identify clusters of individuals based on their daily activity-travel behaviors. Subsequently, the association of activity-travel behaviors and four influence factors (generational cohorts, gender, income, and employment status) is explored by investigating characteristics  within each cluster and contrasting them between clusters. Our contribution is the ability to handle state-of-the-art statistical analysis of large datasets using the divide and combine approach, as well as to construct features that garner topological features of categorical time series.

\subsection{Description of the Activity-Travel Data} \label{Data}

The data for this study was obtained by combining multiple waves of the National Household Travel Survey (NHTS) /National Personal Travel Survey (NPTS). More specifically, the 2001, 2009 and 2017 waves of the NHTS and 1990, and 1995 waves of the NPTS were combined. Each wave of the NHTS/NHPS dataset provides information about the daily activity-travel behaviors of a nationally representative sample. The survey has been sponsored by the Federal Highway Administration and conducted periodically since 1969. 

Datasets are currently available for 1983, 1990, 1995, 2001, 2009 and 2017 and we only used the datasets from five waves of NHTS/NPTS including 1990, 1995, 2001, 2009 and 2017. The 1983 survey was excluded due to data quality issues. 

The surveys asked each sampled participant to report 
all trips he/she made during a designated 24-hour time period, from 4 a.m. of one day until 4 a.m. of the next day, yielding a time series of length $T= 1440$ minutes  per respondent. 
Table \ref{tabsurvey} shows some basic information about this data. Column 1 shows the name of the survey while Column 2 shows the number of available respondents under each survey. For our analysis, we focus on adults (i.e., 18 years or older) who reported their activity-travel on a \textit{typical} weekday (Tuesday, Wednesday, or Thursday), and their counts are shown in Column 3 of the table.
The number of respondents across all surveys for our analysis is $N=250882$. In addition to the activity-travel behavior information, socioeconomic and demographic information of the respondents (i.e., age, gender, employment status, etc.) are also provided for each survey.

\begin{table} 
\caption{Data Sources and Sample Sizes}
\label{tabsurvey}
\centering
\begin{tabular}{|c|c|c|}
\hline 
Data Source & Full Survey & Selected Adults\\ 
\hline
1990 NHTS   & 48385      & 9769                                       \\\hline
1995 NHTS   & 95360      & 20997                                      \\\hline
2001 NHTS   & 160758     & 44201                                      \\\hline
2009 NHTS   & 308901     & 84366                                      \\\hline
2017 NHTS   & 264234     & 91549                                      \\\hline
Total       & 877638     & 250882   \\
\hline                                
\end{tabular}
\end{table}

We denote $N_w$ as the number of participants in survey wave $w$ for $w=1,\ldots, W (=5)$. Then, 
$N=\sum_{w=1}^{W} N_w$. Rather than counting each participant once, we will follow NHTS and assign a ``weight'' $w_n$ to the $n$th participant, $n=1, \ldots,N$. 
The weighting scheme is used in order to produce valid population-level estimates by trying to reduce nonresponse bias and sampling bias. This procedure is standard in the analysis of household surveys, including steps of calculating base weights, adjusting the base weights for eligibility and nonresponse, and further poststratifying the adjusted weights to external source data \citep{NHTS2017}; see Table \ref{tab:wts}. The $0$ entries in the table indicate no observations. Specifically, there are no Millennials in Waves 1 and 2 because they were not adults at that time yet. There is no Government Issue Generation in Wave 5 as well.

Different generations are defined based on people's birth year:  Government Issue (GI) Generation (birth year 1901 to 1924); Silent Generation (birth year 1925 to 1943); Baby Boomers (birth year 1944 to 1964); Generation X (birth year 1965 to 1981); Millennials (birth year 1982 to 2000).

\begin{table*}
\caption{Total Weights of Different Demographic Variables}
\label{tab:wts}
\centering
\begin{tabular}{|l|l|l|l|l|l|}
\hline
                  & Wave1       & Wave2       & Wave3       & Wave4       & Wave5       \\ \hline
GI                 & 4101984     & 3607945     & 2993436     & 900313      & 0           \\ \hline
Silence Generation & 10805726    & 10766706    & 12304352    & 9329861     & 5895241     \\ \hline
Baby Boomer        & 22337829    & 23282036    & 27896189    & 27303881    & 28444900    \\ \hline
Generation X       & 7885177     & 13484379    & 24599990    & 25747942    & 26858832    \\ \hline
Millennial         & 0           & 0           & 2614342     & 12573553    & 29109232    \\ \hline
Worker             & 33352378.6  & 37299955.77 & 52174455.13 & 53247605.7  & 61458579.48 \\ \hline
Non-worker         & 11778337.55 & 13841109.19 & 18232436.39 & 22593569.93 & 28846987.6  \\ \hline
Male               & 22772337        & 25938351       & 34993077       & 37756891       & 44694301       \\ \hline
Female             & 22358379        & 25202714       & 35415231       & 38098659       & 45558163    \\
\hline  
\end{tabular}
\end{table*}

\subsection{Study Design}\label{SDn}

We use three activity-travel types to characterize an individual's daily pattern. These include (a) in-home activity, (b) out-of-home activity, and (c) travel. This information is derived by consolidating detailed trip purpose categories provided by the survey.
For each respondent $n=1,\ldots,N(=250882)$ and for each minute $t=1,\ldots,T (=1440)$, we define the categorical time series with $J=3$ levels as follows:
\begin{equation}
x_{n,t} = 
\begin{cases}
0   &  \mbox{if respondent is at Home}  \\
1   &  \mbox{if respondent is on Travel}\\
2   &  \mbox{if respondent is Out of Home}.
\end{cases}
\label{act-trav1}
\end{equation}

\begin{figure*}
\includegraphics[width=\textwidth]{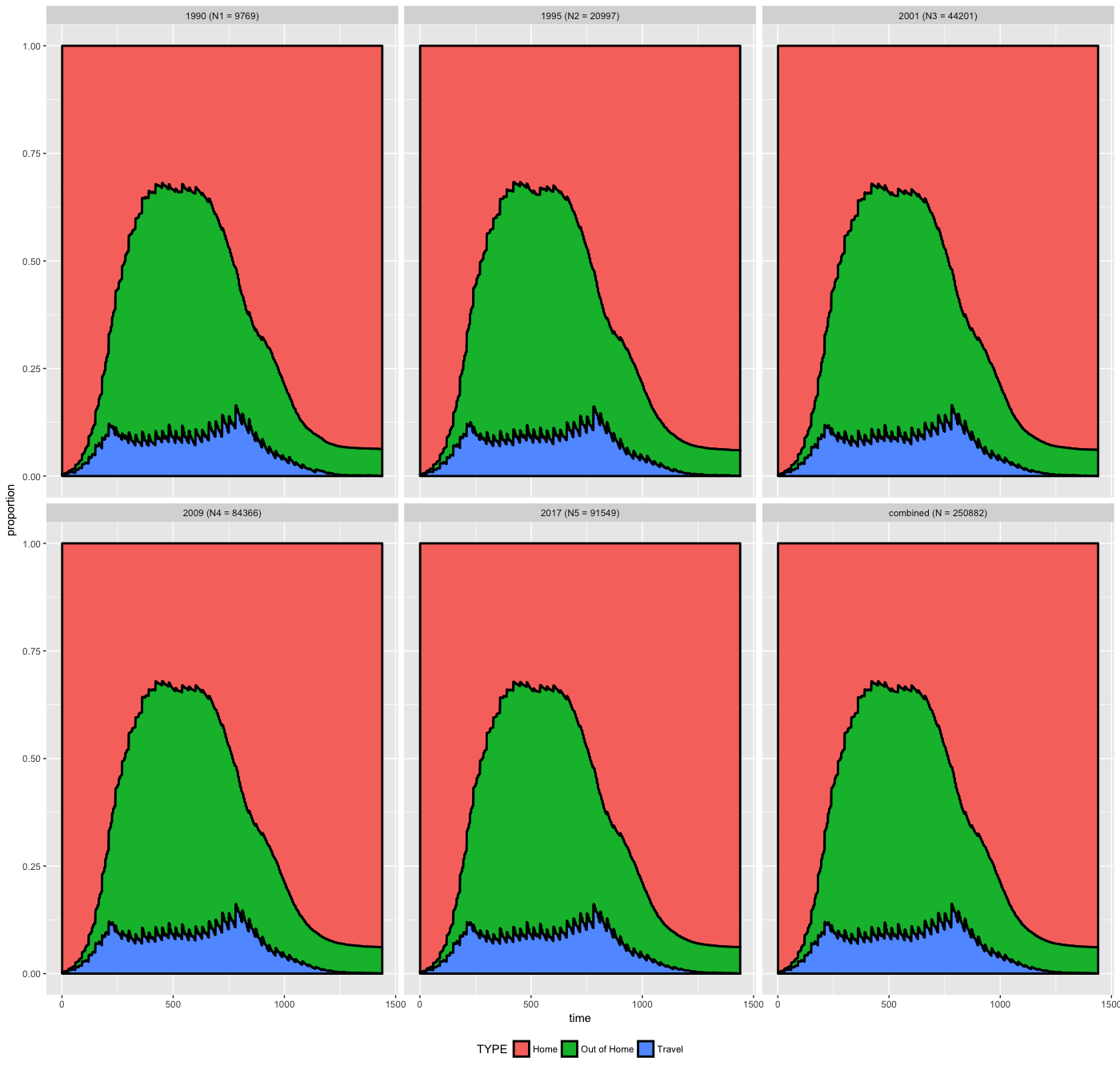}
\caption{Time Course Proportions of the Three Categories for the Five Survey Waves.}
\label{threecategory}
\end{figure*}

Figure \ref{threecategory} shows the proportions of these three categories on the different survey waves. The title for each plot shows the year of the wave and the number of respondents. In general, all waves exhibit similar profiles, with the ``Home'' category having the highest proportion of respondents in the beginning and the end, while the ``Out of Home'' category is dominant during the middle of the day. 

\begin{figure*} 
\centering
\includegraphics[width=\textwidth]{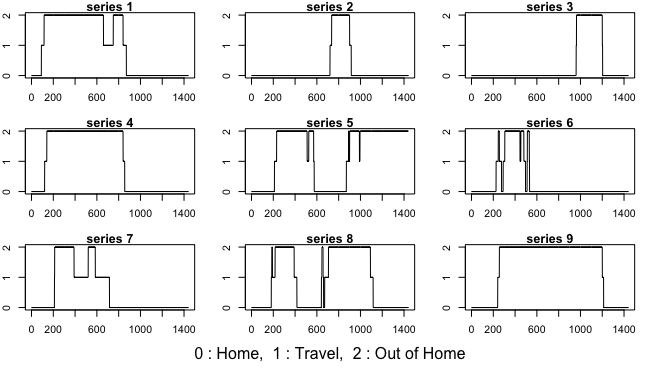}
\caption{Categorical Time Series for Randomly Selected Respondents.}
\label{rawdata}
\end{figure*}

Figure \ref{rawdata} shows the categorical time series for nine randomly selected respondents. The x-axis shows the time in minutes from 4 am on a given day until 4 am of the next day, for a total of $T=1440$ minutes. The y-axis shows in which of the three categories the respondent is at each minute $t$. The figure shows that 
several respondents have normal behaviors, i.e., they go out in the early morning ($t=100\sim 200$ is 5:00 to 8:00 am), spend the daytime outside, and return home in the late afternoon ($t=800\sim1000$ is 6:00 to 9:00 pm). There is another kind of activity-travel pattern where people stay at home most of the time, except for a couple of hours during the afternoon ($700\sim900$ is 3:00 to 7:00 pm).

\subsection{Clustering Respondents by the Divide and Combine Scheme} \label{respclust}

We employ the divide and combine scheme described in Sections \ref{features} and \ref{DCS}. 
We use Step 3.1 to divide the $N(=250882)$ respondents into $S=100$ sets. The first $99$ sets have $2508$ respondents each, while the last set has $2590$ respondents.
Each set is assigned to a different processor on the UConn cluster, as described in Section \ref{DCS}.
Within each of the $S=100$ processors, we extract the first-order persistence landscape corresponding to the WFT of each series. 

For a given number of clusters $K$, we carry out the K-means algorithm in parallel on the $S$ processors (see Step 3.3), in interaction between these processors and the main processor. We then combine the results (see Step 3.4) to arrive at the final stage of clustering the respondents into $K$ groups.

In practice, the number of clusters $K$ is unknown. To select $K$, we use WCSS, a measure of overfitting defined in equation (\ref{WCSS}). 
Table \ref{compareT} shows the values of WCSS and computation times for each value of $K$ ranging from $2$ to $5$. We separate the time cost for the feature extraction and K-means via using UConn HPC cluster with $S=100$ nodes/processors.

\begin{table} 
\caption{Model Comparisons for Choosing Number of Clusters $K$: The number of clusters; WCSS; CPU Time seconds (feature extraction + K-means).}\label{compareT}
\centering
\begin{tabular}{|l|l|c|}\hline
No. & WCSS  & seconds (FE + K-means)\\\hline
$K=2$       & 9.2E4       & 3.3+0.8 \\\hline
$K=3$       & 4.5E4       & 3.3+1.09 \\\hline
$K=4$       & 3.4E4     & 3.3+2.82 \\\hline
$K=5$       & 2.7E4      & 3.3+2.5 \\\hline
\end{tabular}
\end{table}

\begin{figure*} 
\centering
\includegraphics[width=\textwidth]{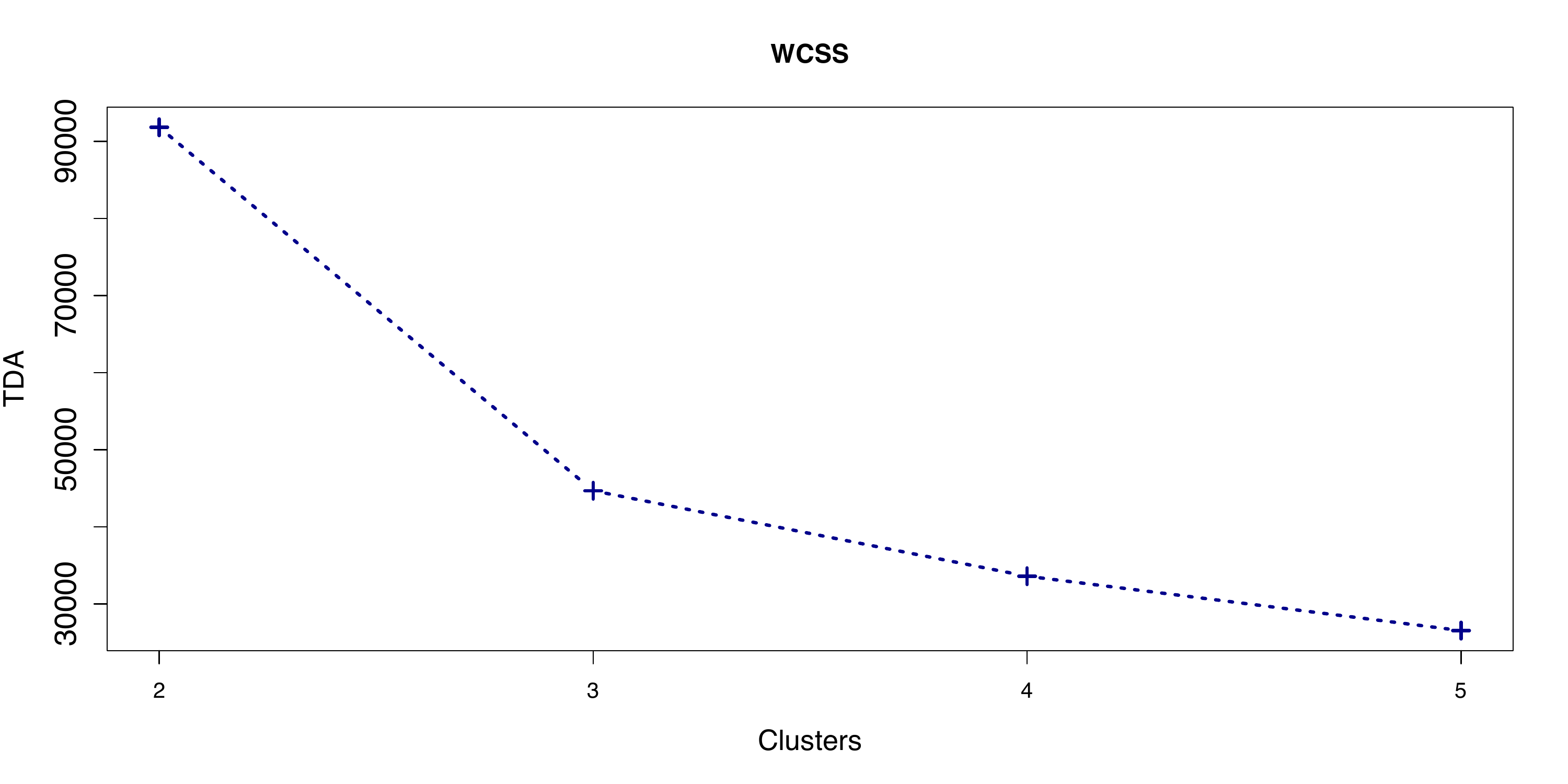}
\caption{WCSS versus Number of Clusters $K$. }
\label{totwithintda}
\end{figure*}

The procedure takes only a few seconds to construct the features and complete the clustering, which indicates that the method is highly computationally effective. .
Figure \ref{totwithintda} plots the WCSS versus the number of clusters $K$. 
Using the Elbow method \citep{Thorndike53whobelongs,ketchen1996application}, we see that the plot selects $K=3$ clusters.

\subsection{Interpretation of Results}\label{interpret}

\begin{figure*}
\centering
\includegraphics[width=\textwidth]{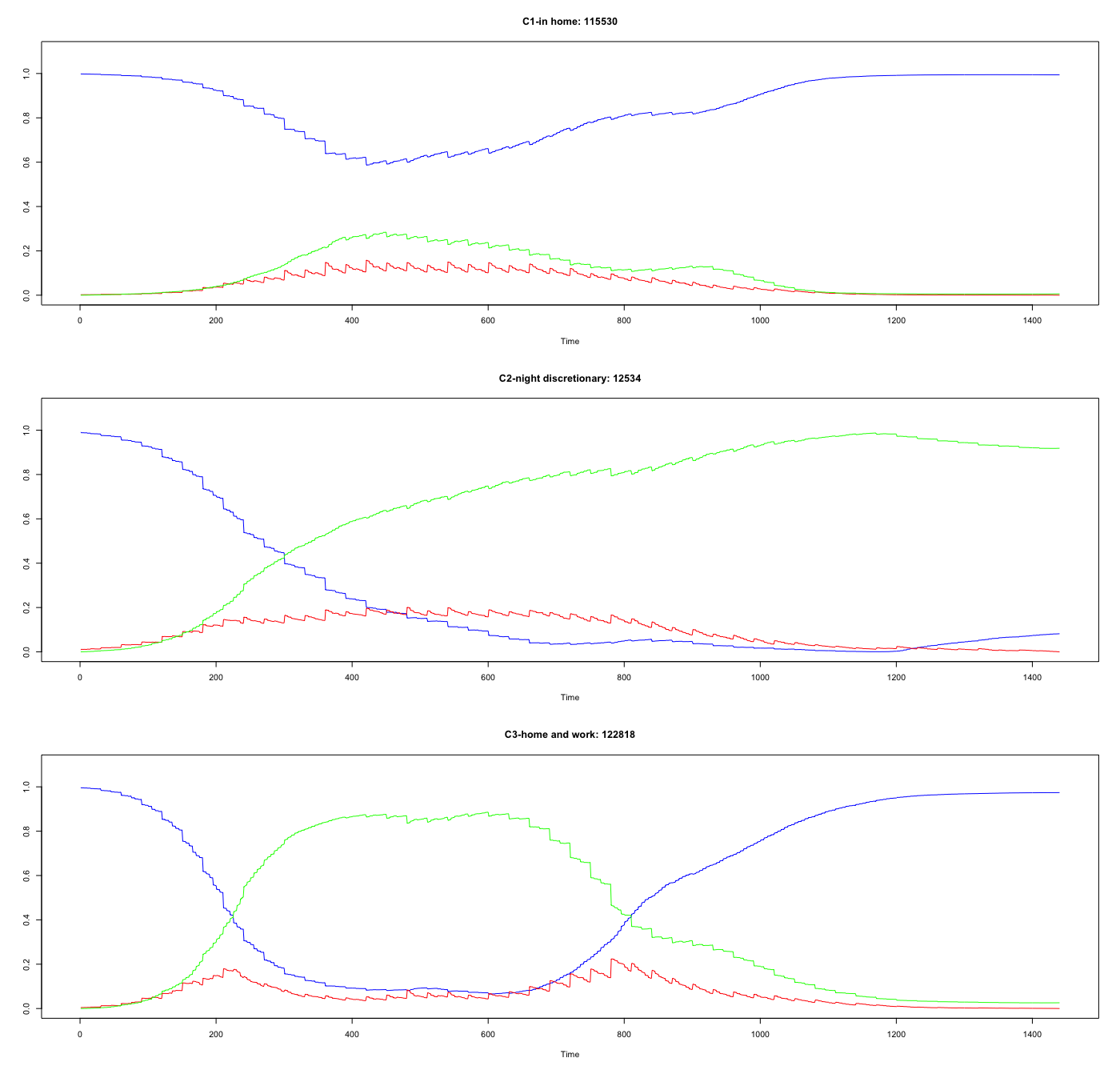}
\caption{Three clusters proportions. The $x$-axis is the minutes and the $y$-axis is the proportion of three categories: Blue-Home, Red-Travel, Green-Out of Home. The title gives the name of the cluster and the size of it. ``C1-in home: 115530'' means that the first proportion plot is the cluster one, called ``in home'' cluster, and there are total 115530 adults in C1.}
\label{3tdakmeans}
\end{figure*}

Figure \ref{3tdakmeans} also shows the proportion of each category over minutes. 
Three clusters were obtained by applying the proposed method. $115530$ ($46.05\%$) respondents fall into cluster 1, $12534$ ($5.00\%$) respondents fall into cluster 2, and $122818$ ($48.95\%$) respondents fall into cluster 3.
Cluster 1 contains adults staying at home most of the time so will be named ``C1-in home''; Cluster 2 is named ``C2-night discretionary'' 
as most of the adults in the cluster would stay in the ``Out of Home'' category until the end of the survey period; 
Cluster 3 is named ``C3-home and work'' as people in the Cluster 3 would stay in the ``Out of Home'' category during the daytime and stay in the ``Home'' category at night.

We are interested in four demographic variables as they are closely related to activity-travel patterns in the literature, generations (GI Generation, Silent Generation, Baby Boomers, Generation X, Millennials), gender (male, female), income (25k-, 25k-55k, 55k-75k, 75k-100k, 100k+), and employment (worker, non-worker). 
In the following, we explore the activity-travel patterns of different survey periods by considering these attributes. 

\begin{figure*}
\centering
\includegraphics[width=\textwidth]{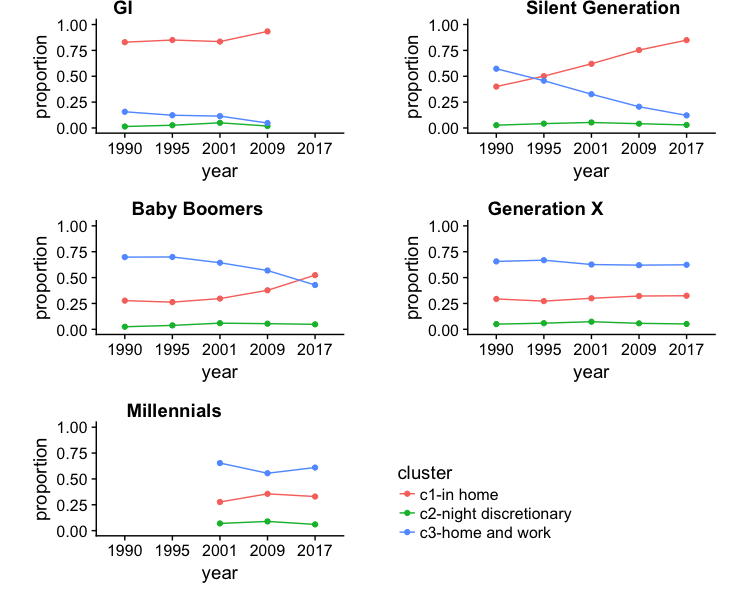}
\caption{The composition of different clusters over five survey periods, as a function of five different generations. 
}
\label{TDAnewgenwaves}
\end{figure*}

In Figure \ref{TDAnewgenwaves}, we can see that (a) most of adults in the GI generation are in ``C1-in home'', which indicates that they are aged; (b) the adults of Silent Generation are moving from ``C3-home and work'' to ``C1-in home'', which can be the sign of them aging, the same as the Baby Boomers; (c) the majority of both of Generation X and Millennials are in cluster ``C3-home and work'', which are workers and students.

We then explore the composition of different clusters over the different survey periods, as functions of demographic variables, like gender, employment and income. 

\begin{figure*}
\centering
\includegraphics[width=\textwidth]{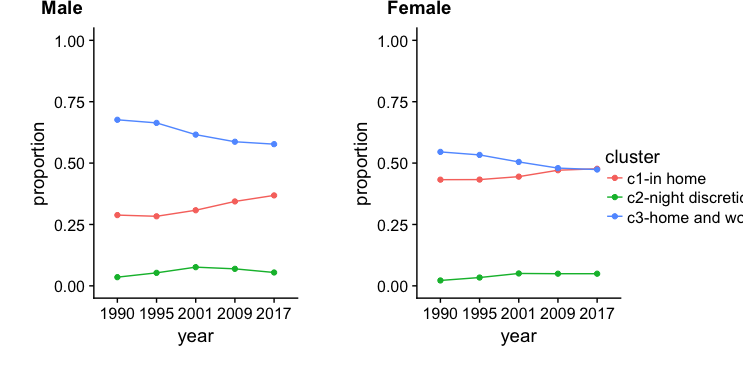}
\caption{The composition of different clusters over five surveys periods, as a function of gender. 
}
\label{TDAnewgenderwaves}
\end{figure*}

In general, Figure \ref{TDAnewgenderwaves} shows that majority of both male and female are in cluster ``C3-home and work'', and the proportions of both of male and femalein cluster ``C1-in home'' increase.  
What is more, starting from 2009, 
the distributions of females in cluster ``C1-in home'' and females in cluster ``C3-home and work'' are about the same, which indicates that there is a trend of female spending more time at home. 

\begin{figure*} 
\centering
\includegraphics[width=\textwidth]{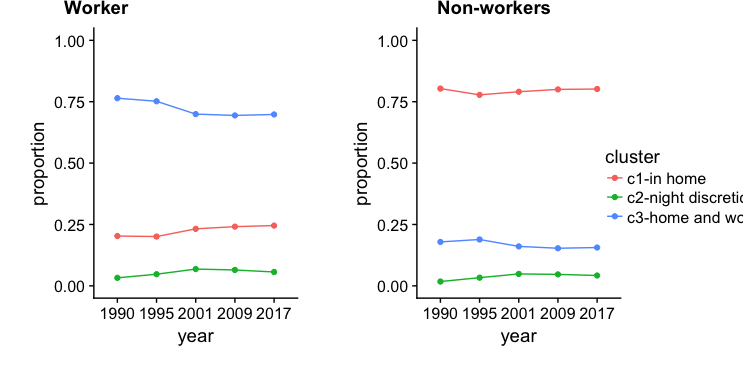}
\caption{The composition of different clusters over five surveys periods, as a function of worker/non-workers.}
\label{TDAnewworkerwaves}
\end{figure*}

Figure \ref{TDAnewworkerwaves} shows a strong connection between the employment types and the clusters. If people are workers, majority of them are in the cluster ``C3-home and work'', and the majority of non-workers are in the cluster ``C1-in home''. On the other hand, it is interesting to see that an increasing trend of workers in the cluster ``C1-in home'' and a decreasing trend of workers in the ``C3-home and work'', which indicates that there are more workers starting to work from home.
      
\begin{figure*}
\centering
\includegraphics[width=\textwidth]{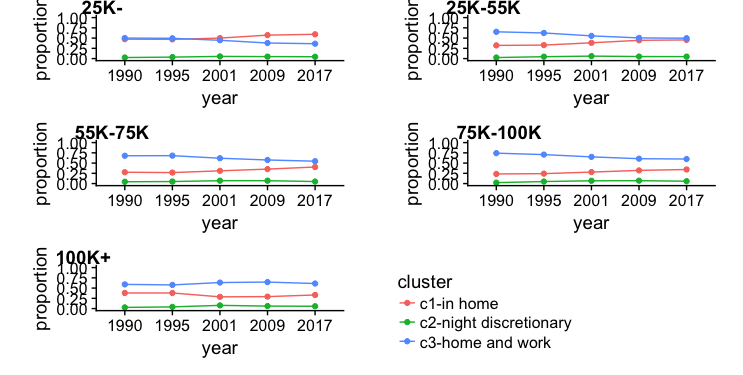}
\caption{The composition of different clusters over five survey periods, as a function of different income levels.}
\label{TDAnewincomewaves}
\end{figure*}

Figure \ref{TDAnewincomewaves} shows the composition for different income levels. It is interesting to see that the middle income levels (from $25K$ to $100K$) have an increasing trend of cluster ``C1-in home'' over years and a decreasing trend of the ``C3-home and work''. Combining with Figure \ref{TDAnewworkerwaves} above, it means that the increasing trend of workers working at home are in the middle income level.

\section{Summary and Discussion} \label{summary}

In order to understand the relationship between individuals' activity-travel behaviors and their demographic characteristics using actively collected ``big'' survey data, a new sequence alignment method to cluster the temporal behaviors is proposed. 
The proposed method is demonstrated using data from NHTS to identify clusters of activity-travel patterns. 
The method uses TDA to construct a first-order persistence landscape which is then used as a feature for clustering. 
The proposed method has been implemented in C++ and the code is posted on Github.

It must be pointed out that there are a large number of other factors that are also highly related to daily activity-travel behaviors, such as, age, life cycle, built environment, etc. however, given the methodology focus of this study, a more comprehensive investigation is left to a follow up paper.

Last but not least, the aggregation procedure of converting features is only focused on the first-order persistence landscape, which is essentially the combination of the maximum and minimum of the Walsh-Fourier Transforms. It is an appropriate approach when the raw time series is relatively simple, not containing too many significant patterns. If the activity-travel patterns are more complex, like a salesman's business day, it could be meaningful to construct higher order persistence landscapes, which will be related to a set of local maxima and minima of the Walsh-Fourier Transforms. 
This will be the subject of future research.

\section*{Appendix: TDA and the First-order Persistence Landscape}

We start with a brief review of Topological Data Analysis (TDA), which
is now an emerging area for analyzing big data with complex structures. Using computational homology, TDA is aimed at analyzing the topological features of data and representing these features using low dimensional representations \citep{CarlssonBulletin}. The input to TDA is often a set of data points (point cloud) or a function, and persistence homology distills essential topological features in the data, which can then be used together with suitable dissimilarity measures to identify patterns in the data sets. We discuss TDA on functions, which is the approach developed in Sections \ref{features} and \ref{DCS}.

\subsection*{Computational Procedure for TDA on Functions}

\begin{figure*}
\centering
\includegraphics[width=\textwidth]{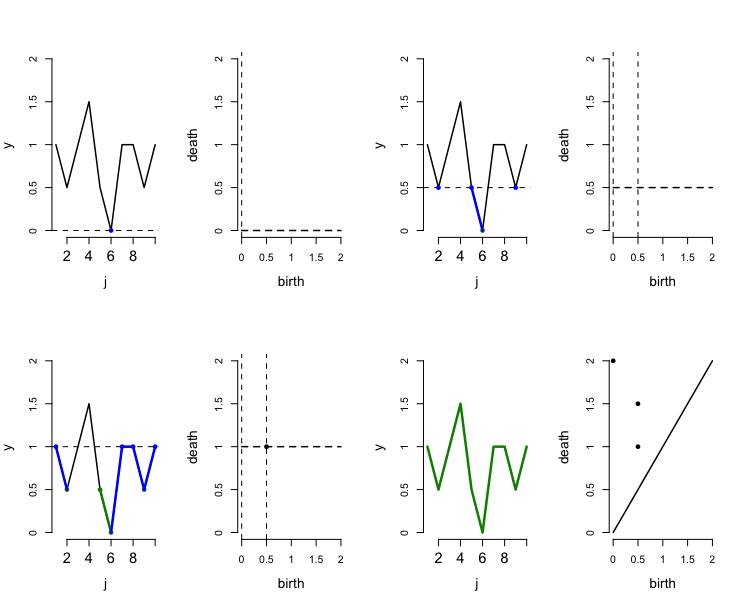}
\caption{Four pairs of plots, in order from top left to bottom right, to illustrate the procedure of getting the persistence diagram on a function}
\label{functionpersis}
\end{figure*}

We look at the method to construct persistence diagrams on functions by using the sublevel set filtration. 
Figure \ref{functionpersis} shows the simple procedure of extracting a persistence diagram from a function. Suppose $y_j = f(j), j=1,\dots, 10$ and let the sublevel set be $L_r = \{y_j|y_j \leq r\}$. TDA is used to construct the persistence diagram based on $L_r$.

\begin{itemize}
\item [(i)] When $r = 0$, a connected component is identified (marked as a blue dot, which is the oldest connected component). The vertical slash line of the second plot records the ``birth time $= 0$'' and the horizontal slash line indicates $r$. There is no point on the birth/death plot, since no connected components died at $r=0$. 

\item [(ii)] When $r = 0.5$, there are two more connected components coming out (indicated in blue); the blue dot in the middle with a blue line connecting it to the dark green dot indicates that the oldest connected component ``enlarges'' and is ``still alive''. The other black vertical slash line in the second plot gives the ``birth time'' for the other two new connected components. There is  no connected component dead yet, and hence no points are shown on the birth/death plot.  

\item[(iii)] When $r = 1$, all old components ``enlarge'' and there is one newer component ``killed'' by the older one. Therefore, there is a ``black dot with birth $= 0.5$ and death $= 1$'' shown on the second plot.

\item[(iv)] When $r = 2$,  the last component is  ``killed, birth $= 0$, death $= 2$'', which is the black dot on the location $(0,2)$. The other black dot corresponding to $(0.5, 1.5)$ of the second plot tells the ``birth and death'' of another connected component.

\end{itemize}

\subsection*{First-Order Persistence Landscape}

First, in the persistence diagram obtained by using the sublevel set filtration, the furthest point away from the diagonal line is always born at the minimum value of the function and dies at the maximum value of the function.

Second, referring to the definition of persistence landscape in Section 2.3 from \cite{Bubenik2015}, given a persistence diagram $\{(b_i, d_i), \forall i\}$, the first-order persistence landscape is 
$$
\mbox{PL}(\ell) = \max_i\{ \min (\ell-b_i, d_i-\ell)_+ \}, 
$$
where  $ \ell$ is a real number.  
Because the persistence diagram uses a sublevel set filtration, it has the point $(d_{\min}, d_{\max})$. 
For all $(b_i, d_i)$ that belong to the persistence diagram,  $d_{\min}\leq b_i \leq d_i \leq d_{\max}$. Therefore,
for any real number $\ell$,
$\ell-d_{\min} \geq \ell-b_i$ and $d_{\max} - \ell \geq d_i - \ell$, which implies that 
$$
\min(\ell-d_{\min}, d_{\max}-\ell)_+ \geq \min(\ell-b_i, d_i-\ell)_+ ,
$$ 
which in turn  implies that
\begin{eqnarray*}
\mbox{PL}(\ell) = \max_i\{ \min (\ell-b_i, d_i-\ell)_+ \} \nonumber\\
= \min(\ell-d_{\min}, d_{\max}-\ell)_+.
\end{eqnarray*}

Finally, let $(d_{\min}, d_{\max}) \subset (D_{\min}, D_{\max})$ and taking grids   
$\{D_{\min}+\frac{(\ell-1)*(D_{\max}-D_{\min})}{L-1},\ell=1, 2, \ldots  , L\}$, we have 
$$
\mbox{PL}(\ell) = \min (V_1 (\ell), V_2 (\ell))_{+},
$$
where,
\begin{eqnarray*}
V_1(\ell) = D_{\min} + \frac{(\ell-1) (D_{\max}-D_{\min})}{L-1} - d_{\min} \nonumber\\
V_2(\ell) = d_{\max} - D_{\min} - \frac{(\ell-1) (D_{\max} - D_{\min})}{L-1}.
\end{eqnarray*}
These expressions will be used on the WFT function obtained from  each time series $n=1,\ldots,N$ in Section \ref{features}.


%
%
\section*{Conflict of interest}
The authors declare that they have no conflict of interest.


\begin{thebibliography}{1}

\bibitem[{Bubenik(2015)}]{Bubenik2015}
Bubenik P (2015) Statistical topological data analysis using persistence
  landscapes. J Mach Learn Res 16(1):77--102

\bibitem[{Calabrese et~al.(2013)Calabrese, Diao, Lorenzo, Ferreira, and
  Ratti}]{CALABRESE2013301}
Calabrese F, Diao M, Lorenzo GD, Ferreira J, Ratti C (2013) Understanding
  individual mobility patterns from urban sensing data: A mobile phone trace
  example. Transportation Research Part C: Emerging Technologies 26:301 -- 313

\bibitem[{Candia et~al.(2008)Candia, Gonz{\'{a}}lez, Wang, Schoenharl, Madey,
  and Barab{\'{a}}si}]{Candia_2008}
Candia J, Gonz{\'{a}}lez MC, Wang P, Schoenharl T, Madey G, Barab{\'{a}}si AL
  (2008) Uncovering individual and collective human dynamics from mobile phone
  records. Journal of Physics A: Mathematical and Theoretical 41(22):224015

\bibitem[{Carlsson(2009)}]{CarlssonBulletin}
Carlsson G (2009) {Topology and data}. Bulletin of the American Mathematical
  Society 46(2):255--308

\bibitem[{Edelsbrunner and Harer(2010)}]{harer2010}
Edelsbrunner H, Harer J (2010) {Computational Topology. An Introduction.}
  American Mathematical Society

\bibitem[{Figueiras et~al.(2016)Figueiras, Silva, Ramos, Guerreiro, Costa, and
  Jardim-Goncalves}]{Figueiras2016}
Figueiras P, Silva R, Ramos A, Guerreiro G, Costa R, Jardim-Goncalves R (2016)
  Big data processing and storage framework for its: A case study on dynamic
  tolling. ASME 2016 International Mechanical Engineering Congress and
  Exposition

\bibitem[{Goulias(1999)}]{GOULIAS1999535}
Goulias KG (1999) Longitudinal analysis of activity and travel pattern dynamics
  using generalized mixed markov latent class models. Transportation Research
  Part B: Methodological 33(8):535 -- 558

\bibitem[{Huang et~al.(2018)Huang, Levinson, Wang, Zhou, and Wang}]{Huang12710}
Huang J, Levinson D, Wang J, Zhou J, Wang Zj (2018) Tracking job and housing
  dynamics with smartcard data. Proceedings of the National Academy of Sciences
  115(50):12710--12715

\bibitem[{Jandui~Silva(2015)}]{silva2015}
Jandui~Silva LLVSFF Bárbara~França (2015) Towards smart traffic lights using
  big data to improve urban traffic. SMART 2015: The Fourth International
  Conference on Smart Systems, Devices and Technologies

\bibitem[{Joh et~al.(2001)Joh, Arentze, and Timmermans}]{joh2001pattern}
Joh CH, Arentze T, Timmermans H (2001) Pattern recognition in complex activity
  travel patterns: comparison of euclidean distance, signal-processing
  theoretical, and multidimensional sequence alignment methods. Transportation
  Research Record: Journal of the Transportation Research Board (1752):16--22

\bibitem[{Ketchen and Shook(1996)}]{ketchen1996application}
Ketchen DJ, Shook CL (1996) The application of cluster analysis in strategic
  management research: an analysis and critique. Strategic management journal
  17(6):441--458

\bibitem[{Kwan(2000)}]{Kwan2000}
Kwan MP (2000) Interactive geovisualization of activity-travel patterns using
  three dimensional geographical information systems: a methodological
  exploration with a large data set. Transportation Research Part C: Emerging
  Technologies 8:185--203

\bibitem[{Pas(1988)}]{Pas1988}
Pas EI (1988) Weekly travel-activity behavior. Transportation 15(1):89--109

\bibitem[{Recker et~al.(1985)Recker, McNally, and Root}]{RECKER1985279}
Recker WW, McNally MG, Root GS (1985) Travel/activity analysis: Pattern
  recognition, classification and interpretation. Transportation Research Part
  A: General 19(4):279 -- 296

\bibitem[{Shanks(1969)}]{Shanks1969}
Shanks JL (1969) Computation of the fast walsh-fourier transform. IEEE Trans
  Comput 18(5):457--459

\bibitem[{Shelley Brock~Roth(2017)}]{NHTS2017}
Shelley Brock~Roth JD Yiting~Dai (2017) 2017 nhts weighting report. National
  Household Travel Survey

\bibitem[{Shoval and Isaacson(2007)}]{Shoval2007}
Shoval N, Isaacson M (2007) Sequence alignment as a method for human activity
  analysis in space and time. Annals of the Association of American Geographers
  97:282 -- 297

\bibitem[{Stoffer(1991)}]{Stoffer1991}
Stoffer DS (1991) Walsh-fourier analysis and its statistical applications.
  Journal of the American Statistical Association 86(414):461--479

\bibitem[{Stolz et~al.(2017)Stolz, Harrington, and
  Porter}]{stolz2017persistent}
Stolz BJ, Harrington HA, Porter MA (2017) Persistent homology of time-dependent
  functional networks constructed from coupled time series. Chaos: An
  Interdisciplinary Journal of Nonlinear Science 27(4):047410

\bibitem[{Thorndike(1953)}]{Thorndike53whobelongs}
Thorndike RL (1953) Who belongs in the family. Psychometrika pp 267--276

\bibitem[{Wang et~al.(2018)Wang, Ombao, and Chung}]{wang2018topological}
Wang Y, Ombao H, Chung MK (2018) Topological data analysis of single-trial
  electroencephalographic signals. The annals of applied statistics 12(3):1506

\bibitem[{Wilson(2001)}]{Wilson2001}
Wilson C (2001) Activity patterns of canadian women: Application of clustalg
  sequence alignment software. Transportation Research Record 1777(1):55--67

\bibitem[{Zhang et~al.(2018)Zhang, Kang, Axhausen, and Kwon}]{ZHANG201896}
Zhang A, Kang JE, Axhausen K, Kwon C (2018) Multi-day activity-travel pattern
  sampling based on single-day data. Transportation Research Part C: Emerging
  Technologies 89:96 -- 112
\end{thebibliography}

\end{document}